\documentclass{Interspeech2024}
\usepackage{amsmath,graphicx}
\usepackage{lineno,hyperref}
\usepackage{algorithm}
\usepackage{algpseudocode}
\usepackage{graphicx}
\usepackage[table,xcdraw]{xcolor}
\usepackage[justification=centering]{caption}
\usepackage{caption}
\usepackage{multirow}
\usepackage{etoolbox}
\usepackage{hyperref}
\usepackage{array}
\usepackage{subcaption}
\usepackage{ amssymb }
\usepackage{multirow}
\usepackage{comment}
\usepackage{subcaption}
\usepackage{placeins}
\usepackage{bm}
\usepackage{listings}
\usepackage{xcolor}

\definecolor{codegreen}{rgb}{0.0, 0.5, 0.0}
\definecolor{codegray}{rgb}{0.95,0.5,0.5}
\definecolor{codepurple}{rgb}{0.58,0,0.82}
\definecolor{backcolour}{rgb}{1, 1, 1}
\definecolor{Darkgreen}{rgb}{0.0, 0, 1}
\interspeechcameraready

\title{TM-PATHVQA: 90000+ Textless Multilingual Questions for Medical Visual Question Answering}

\name{Tonmoy}{Rajkhowa}
\name{Amartya}{Roy Chowdhury}
\name{Sankalp}{Nagaonkar}
\name{Achyut}{Mani Tripathi}
\name{S R Mahadeva}{Prasanna}

\address{
  Indian Institute of Technology, Dharwad, India}
\email{212022001@iitdh.ac.in, amartya.chowdhury@iitdh.ac.in, 210020031@iitdh.ac.in, t.achyut@iitdh.ac.in, prasanna@iitdh.ac.in}


\newcommand{\red}[1]{\textcolor{blue}{#1}}

\begin{document}

\maketitle

\begin{abstract}

In healthcare and medical diagnostics, Visual Question Answering (VQA) may emerge as a pivotal tool in scenarios where analysis of intricate medical images becomes critical for accurate diagnoses. Current text-based VQA systems limit their utility in scenarios where hands-free interaction and accessibility are crucial while performing tasks. A speech-based VQA system may provide a better means of interaction where information can be accessed while performing tasks simultaneously. To this end, this work implements a speech-based VQA system by introducing a Textless Multilingual Pathological VQA (TM-PathVQA) dataset, an expansion of the PathVQA dataset, containing spoken questions in English, German \& French. This dataset comprises 98,397   multilingual spoken questions and answers based on 5,004 pathological images along with 70 hours of audio. Finally, this work benchmarks and compares TM-PathVQA systems implemented using various combinations of acoustic and visual features.

\end{abstract}

\section{Introduction}

In the realm of healthcare and medical diagnostics, Visual Question Answering (VQA)~\cite{fang2015captions, chen2014learning} may emerge as a pivotal tool for analyzing complex medical images~\cite{8419425}. It may enable healthcare professionals to inquire about specific details within the visuals, fostering a deeper understanding. VQA may bridge complex medical visuals and human interpretation, thereby improving healthcare diagnosis. However, current VQA systems rely on text-based questions~\cite{Antol_2015_ICCV,8578734}, limiting their utility in scenarios where hands-free interaction and accessibility are crucial, particularly in healthcare settings. Hence, integrating speech may enhance the user experience by offering a more natural mode of interaction while performing tasks simultaneously~\cite{zhang2017speech,8929263,10343139,8296600}. Thus, a speech-based VQA system would allow for hands-free operation where typing might be cumbersome. Hence, with this motivation, this work proposes a spoken VQA system that can accept multilingual spoken queries. The responses could still be displayed in textual form, enabling better perception and documentation for future references. \par

The development of a clinically significant spoken VQA system requires training using a dataset comprising spoken questions along with textual answers based on medical images. Without any such existing dataset, the PathVQA dataset~\cite{he2020pathvqa} is extended by converting the textual questions into spoken form using a Direct Text-to-Speech Translation (DT2ST) ~\cite{barrault2023seamlessm4t} system. Subsequently, this resulted in the creation of a Textless Multilingual Pathological Visual Question Answering (TM-PathVQA) dataset, featuring spoken questions in English, German and French, tailored for this task. Notably, this paper also implements a Multi-Modal Learning (MML) framework to evaluate the efficacy of TM-PathVQA dataset by addressing the "Yes or No" and open-ended type questions separately. Furthermore, this work also presents a comparative analysis between the frameworks implemented by incorporating various combinations of audio and image features extracted using various state-of-the-art models. \par 

Development of VQA systems for the medical domain began with the introduction of the VQA-Med~\cite{ImageCLEFmedicalCaptionOverview2023} dataset, which contains medical images and questions that require analysis of both visual content and medical context. VQA-RAD~\cite{zhan2020medical} incorporates Radiology reports into the VQA task for medical images. However, these two datasets were domain-specific. This led to the creation of PathVQA~\cite{he2020pathvqa} covering diverse Pathological contents. Hence, these attributes motivated us to extend the PathVQA to contain spoken questions. To the best of our knowledge, this is the first-ever VQA dataset that incorporates speeches to facilitate the development of a spoken VQA system. \par

This paper presents three primary contributions. Firstly, it introduces the first-ever TM-PathVQA dataset featuring spoken questions in four languages \textit{viz.} English, German, and French.
Secondly, it also introduces a novel framework for implementing TM-PathVQA systems. Finally, this paper presents a diverse set of benchmark evaluations on this dataset by comparing TM-PathVQA systems that were implemented using various audio and image features extracted using various state-of-the-art models. The overview of this paper is as follows: Section 2 provides an overview of the TM-PathVQA dataset and its development process. Section 3 describes the details of different Multimodal Learning Frameworks (MML). Section 4 presents the benchmark results and discussions. Finally, Section 5 concludes the paper and suggests potential directions for future research.

\begin{figure*}[htbp!]
	\centering
		\begin{subfigure}[h]{0.12\textwidth}
			\includegraphics[width=2.5cm, height=2.3cm]{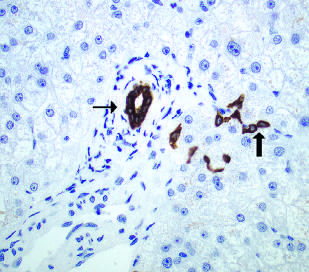}
			\caption{}
			\label{AF7}
		\end{subfigure}~~~~~~~~~
		\begin{subfigure}[h]{0.20\textwidth}
			\includegraphics[width=3.5cm, height=2.3cm]{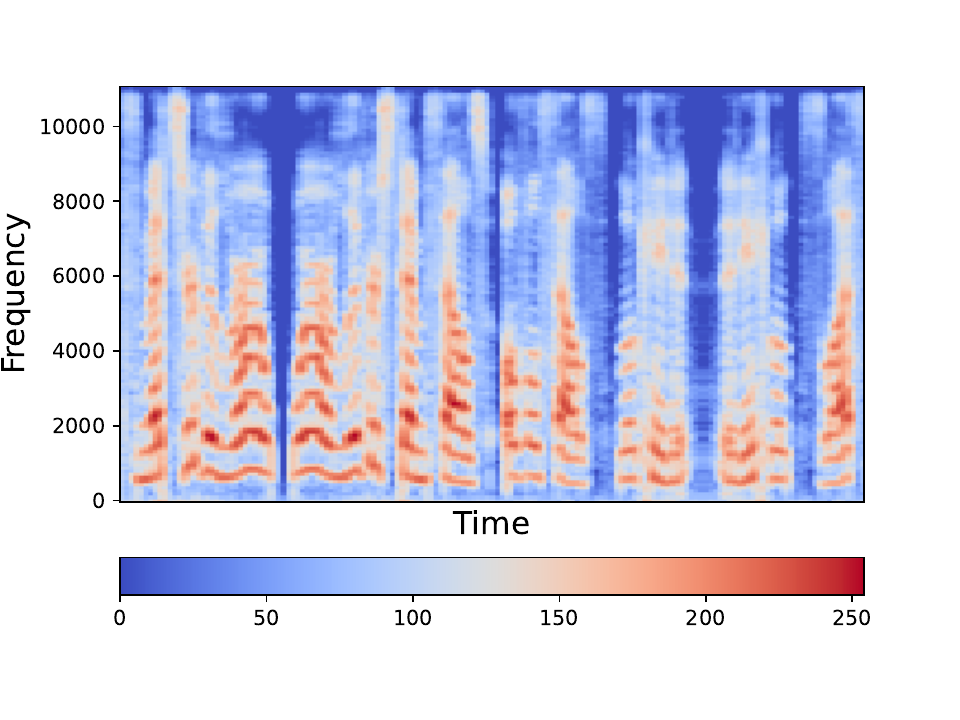}
			\caption{}
			\label{AF8}
		\end{subfigure} %\qquad
		\begin{subfigure}[h]{0.20\textwidth}
			\includegraphics[width=3.5cm, height=2.3cm]{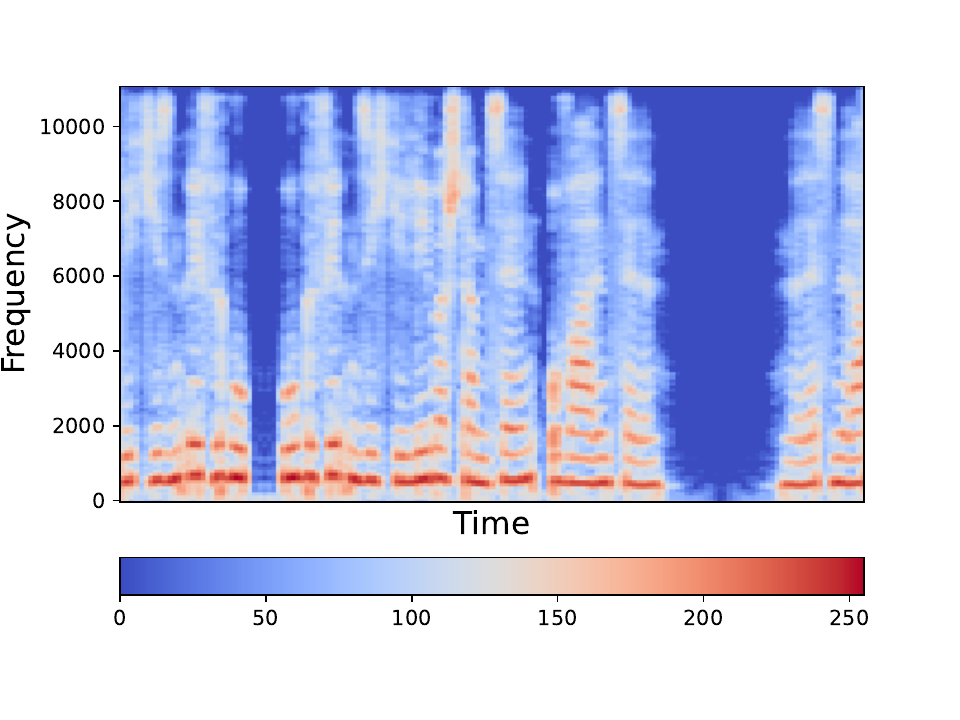}
			\caption{}
			\label{AF1}
		\end{subfigure} %\qquad
			\begin{subfigure}[h]{0.20\textwidth}
			\includegraphics[width=3.5cm, height=2.3cm]{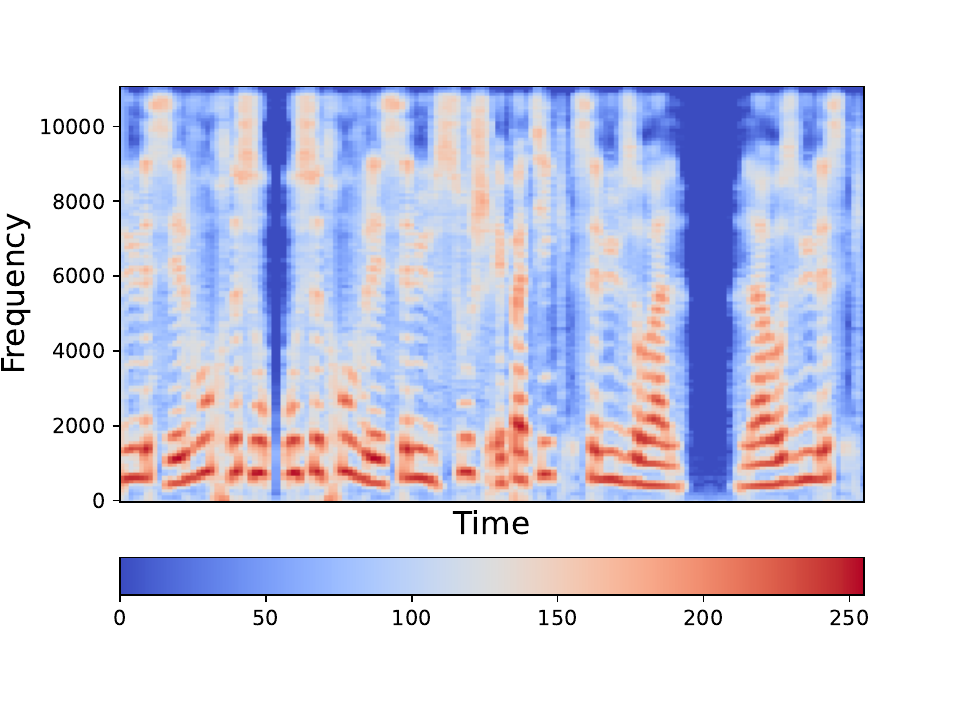}
			\caption{}
			\label{AF2}
		\end{subfigure} %\qquad
\caption{(a) Pathological Image and Spectrogram Visualization of Corresponding Speech-Based Question \& Answer in (b) English (\textbf{Q:} \textcolor{Darkgreen}{" Where are liver stem cells (oval cells) located?"}, \textbf{Ans:} \textcolor{red}{In the canals of hering.}), (c) German (\textbf{Q:} \textcolor{Darkgreen}{" Wo befinden sich Leberstammzellen (ovale Zellen)?"}, \textbf{Ans:} \textcolor{red}{In the canals of hering.}) and (d) French (\textbf{Q:} \textcolor{Darkgreen}{" Où se trouvent les cellules souches hépatiques (cellules ovales)?"}, \textbf{Ans:} \textcolor{red}{In the canals of hering.}) Languages from the TM-PathVQA Dataset.}\label{TAQA_Spectropgram}
\end{figure*}

\section{TM-PathVQA Dataset}

The TM-PathVQA dataset is created by extending the PathVQA dataset. PathVQA is recognized as the most extensive dataset developed for pathological VQA tasks. The textual questions in PathVQA were converted into English, German, and French speeches using the SeamlessM4T multimodal and multilingual AI translation model~\cite{barrault2023seamlessm4t}. This resulted in a dataset containing medical visuals paired with spoken questions in three languages \textit{viz.} English, German, and French, alongside answers in English text. This led to the inclusion of 98,397 question-answer pairs (32,799 questions for each language $\times$ 3 =  98,397 multilingual questions) derived from 5,004 pathological images along with 70 hours of audio containing spoken questions in those three languages. An illustration in Figure \ref{TAQA_Spectropgram} demonstrates a sample including a medical image and corresponding spectrogram visualizations of English, German, and French questions, along with their answers. Subsequently, the TM-PathVQA dataset is partitioned into the non-overlapping train, validation, and test sets to cover both "Yes / No" and open-ended type questions. A comparative statistical overview of the TM-PathVQA dataset is presented in Table \ref{tab:PathVQA-statistics}, with the final dataset containing 98,397 multilingual question-answer pairs associated with 5,004 medical images.
% \usepackage{multirow}
% \usepackage{multirow}
\begin{table}[!hbtp]
\centering
\caption{Specification of TM-PathVQA dataset}
\label{tab:PathVQA-statistics}
\scalebox{0.70}{
\begin{tabular}{c|c|c|ccc}
\hline
\multirow{2}{*}{\textbf{Set}} & \multirow{2}{*}{\textbf{\# of Images}} & \multirow{2}{*}{\textbf{\# of QA pairs}} & \multicolumn{3}{c}{\textbf{Audio duration (in hours)}}                                        \\ \cline{4-6} 
                              &                                        &                                          & \multicolumn{1}{c|}{\textbf{\textcolor{codegreen}{ENGLISH}}} & \multicolumn{1}{c|}{\textbf{\textcolor{codegreen}{GERMAN}}} & \textbf{\textcolor{codegreen}{FRENCH}} \\ \hline
Train                         & 3,021                                  & 19,755                                   & \multicolumn{1}{c|}{12.87}            & \multicolumn{1}{c|}{15.25}           & 13.89           \\ %\hline
Validation                    & 992                                    & 6,279                                    & \multicolumn{1}{c|}{4.03}             & \multicolumn{1}{c|}{4.79}            & 4.34            \\ %\hline
Test                          & 991                                    & 6,761                                    & \multicolumn{1}{c|}{4.4}              & \multicolumn{1}{c|}{5.23}            & 4.75            \\ \hline
\textbf{Total}                & \textbf{5,004}                         & \textbf{32,799}                          & \multicolumn{1}{c|}{\textbf{21.3}}    & \multicolumn{1}{c|}{\textbf{25.27}}  & \textbf{22.98}  \\ \hline
\end{tabular}}
\end{table}
\section{Experimental Methodology}
This section outlines the implementation of TM-PathVQA system in details.

% \usepackage{multirow}
\begin{table*}[!hbt]
\centering
\caption{Performances of English text-based TM-PathVQA systems for Multiclass Classification}
\begin{tabular}{c|c|c|ccccc}
\hline 
\multirow{2}{*}{\textbf{Sl. No.}} & \multirow{2}{*}{\textbf{Text Features}} & \multirow{2}{*}{\textbf{Image Features}} & \multicolumn{5}{c}{\textbf{Performances expressed in \%}}                                                                                                                           \\ \cline{4-8} 
                                  &                                         &                                          & \multicolumn{1}{c|}{\textbf{Top-1}} & \multicolumn{1}{c|}{\textbf{Bleu-1}} & \multicolumn{1}{c|}{\textbf{Bleu-2}} & \multicolumn{1}{c|}{\textbf{Bleu-3}} & \textbf{F1-score} \\ \hline  
1                                 & LaBSE                                   & ViT                       & \multicolumn{1}{c|}{45.29 \textcolor{blue}{$\downarrow$ (6.5)}}                       & \multicolumn{1}{c|}{45.53 \textcolor{blue}{$\downarrow$ (6.84)}}                & \multicolumn{1}{c|}{2.27 \textcolor{blue}{$\downarrow$ (1.63)}}                & \multicolumn{1}{c|}{0.19 \textcolor{blue}{$\downarrow$ (0.27)}}                &   2.24 \textcolor{blue}{$\downarrow$ (2.14)}               \\ %\hline
2                                 & LaBSE                                   & ResNet-152                               & \multicolumn{1}{c|}{43.81 \textcolor{blue}{$\downarrow$ (7.98)}}                       & \multicolumn{1}{c|}{44.81 \textcolor{blue}{$\downarrow$ (7.56)}}                & \multicolumn{1}{c|}{1.93 \textcolor{blue}{$\downarrow$ (1.97)}}                & \multicolumn{1}{c|}{0 \textcolor{blue}{$\downarrow$ (0.46)}}                &     1.64  \textcolor{blue}{$\downarrow$ (2.74)}            \\ %\hline
3                                 & LaBSE                                   & VGG 19                                   & \multicolumn{1}{c|}{47.69 \textcolor{blue}{$\downarrow$ (4.1)}}                       & \multicolumn{1}{c|}{48.12 \textcolor{blue}{$\downarrow$ (4.25)}}                & \multicolumn{1}{c|}{2.18 \textcolor{blue}{$\downarrow$ (1.72)}}                & \multicolumn{1}{c|}{0.16 \textcolor{blue}{$\downarrow$ (0.3)}}                &       3.36 \textcolor{blue}{$\downarrow$ (1.02)}           \\ %\hline
4                                 & LaBSE                                   & Faster RCNN                              & \multicolumn{1}{c|}{\textbf{51.79}}                       & \multicolumn{1}{c|}{\textbf{52.37}}                & \multicolumn{1}{c|}{\textbf{3.9}}                & \multicolumn{1}{c|}{\textbf{0.46}}                &        \textbf{4.38}           \\ \hline 
\end{tabular}

\label{tab:English-text-multiclass}
\end{table*}

\begin{table*}[!hbt]
\centering
\caption{Performances of multilingual speech-based TM-PathVQA systems for Binary Classification using Top-1 accuracy metric (expressed in \%)}
%\scalebox{0.80}
{
\begin{tabular}{c|c|c|c|c|c}
\hline 
\textbf{Sl. No.} & \textbf{Audio Features} & \textbf{Image Features} & \textbf{\textcolor{codegreen}{ENGLISH}} & \textbf{\textcolor{codegreen}{GERMAN}} & \textbf{\textcolor{codegreen}{FRENCH}} \\ \hline 
1                & Wav2Vec2               & ViT      & 52.38 \textcolor{blue}{$\downarrow$ (4.54)}           & 54.37 \textcolor{blue}{$\downarrow$ (2.39)}          & 52.37    \textcolor{blue}{$\downarrow$ (3.99)}       \\ %\hline
2                & Wav2Vec2               & ResNet-152              & 53.38  \textcolor{blue}{$\downarrow$ (3.54)}          & 53.92   \textcolor{blue}{$\downarrow$ (2.84)}        & 52.92   \textcolor{blue}{$\downarrow$ (3.44)}        \\ %\hline
3                & Wav2Vec2               & VGG 19                  & 54.38 \textcolor{blue}{$\downarrow$ (2.54)}           & 54.38  \textcolor{blue}{$\downarrow$ (2.38)}         & 52.38  \textcolor{blue}{$\downarrow$ (3.98)}         \\ %\hline
4                & Wav2Vec2               & Faster RCNN             & \textbf{56.92}   & \textbf{56.76}  & \textbf{56.36}  \\ \hline
5                & Hu-BERT                & ViT      & 56.12 \textcolor{blue}{$\downarrow$ (2.32)}           & 56.22  \textcolor{blue}{$\downarrow$ (1.09)}         & 55.46  \textcolor{blue}{$\downarrow$ (1.97)}         \\ %\hline
6                & Hu-BERT                & ResNet-152              & 56.26  \textcolor{blue}{$\downarrow$ (2.18)}          & 56.67  \textcolor{blue}{$\downarrow$ (0.64)}         & 55.65  \textcolor{blue}{$\downarrow$ (1.78)}         \\ %\hline
7                & Hu-BERT                & VGG 19                  & 56.85  \textcolor{blue}{$\downarrow$ (1.59)}          & 54.78  \textcolor{blue}{$\downarrow$ (2.53)}         & 55.77 \textcolor{blue}{$\downarrow$ (1.66)}          \\ %\hline
8                & Hu-BERT                & Faster RCNN             & \textbf{58.44}   & \textbf{57.31}  & \textbf{57.43}  \\ \hline
9                & melfilterbank          & ViT      & 53.77 \textcolor{blue}{$\downarrow$ (3.08)}           & 51.77 \textcolor{blue}{$\downarrow$ (3.57)}          & 50.23   \textcolor{blue}{$\downarrow$ (6.04)}        \\ %\hline
10               & melfilterbank          & ResNet-152              & 54.53 \textcolor{blue}{$\downarrow$ (2.32)}           & 53.65 \textcolor{blue}{$\downarrow$ (1.69)}          & 52.44 \textcolor{blue}{$\downarrow$ (3.83)}          \\ %\hline
11               & melfilterbank          & VGG 19                  & 54.73 \textcolor{blue}{$\downarrow$ (2.12)}           & 50.28 \textcolor{blue}{$\downarrow$ (5.06)}          & 50.96 \textcolor{blue}{$\downarrow$ (5.31)}          \\ %\hline
12               & melfilterbank          & Faster RCNN             & \textbf{56.85}   & \textbf{55.34}  & \textbf{56.27}  \\ \hline
13               & Whisper                & ViT      & 50.24 \textcolor{blue}{$\downarrow$ (1.4)}           & 50.67 \textcolor{blue}{$\downarrow$ (0.87)}           & 50.16  \textcolor{blue}{$\downarrow$ (1.05)}         \\ %\hline
14               & Whisper                & ResNet-152              & 51.28  \textcolor{blue}{$\downarrow$ (0.36)}          & 50.01 \textcolor{blue}{$\downarrow$ (1.53)}          & 50.48  \textcolor{blue}{$\downarrow$ (0.73)}         \\ %\hline
15               & Whisper                & VGG 19                  & 51.11 \textcolor{blue}{$\downarrow$ (0.53)}            & 51.03 \textcolor{blue}{$\downarrow$ (0.51)}          & 50.33 \textcolor{blue}{$\downarrow$ (0.88)}          \\ %\hline
16               & Whisper                & Faster RCNN             & \textbf{51.64}   & \textbf{51.54}  & \textbf{51.21}  \\ \hline 
\end{tabular}}

\label{tab:English-audio-binary}
\end{table*}

\subsection{Representations for Different Modalities} Various types of feature representations that were employed to implement the system are as follows:

\subsubsection{\textbf{Text Representation}}
The Language-agnostic BERT Sentence Encoder (LaBSE)~\cite{feng2020language}, a BERT-based model~\cite{devlin2018bert}, was used to generate embeddings from texts. Inspired by the success of LaBSE, this work utilizes its capabilities to extract features from text-based questions.

\subsubsection{\textbf{Audio Representation}}
Audio features were extracted from the raw spoken questions, sampled at 16 $kHz$, using Wav2Vec2~\cite{baevski2020wav2vec} (having 1920 dimensions) and Hu-BERT~\cite{hsu2021hubert} (having 768 dimensions) along with $80$-dimensional Mel filterbanks having a window length of $400$ and a hop length of $160$ frames. For Wav2Vec2 features, the XLS-R $128$~\cite{babu2021xls} pre-trained model was employed. Hu-BERT features were extracted from the $11^{th}$ layer of the encoder and then normalized before feeding into the MML framework. Features were extracted from the last encoder layer using the Whisper large-V3~\cite{radford2023robust} pre-trained model.

\subsubsection{\textbf{Visual Representation}} 
Image features were extracted using state-of-the-art models, including Vision Transformer (ViT)~\cite{dosovitskiy2020image}, ResNet-152~\cite{he2016deep}, VGG19~\cite{simonyan2014very}, and Faster-RCNN~\cite{ren2015faster}. ViT divides the image into overlapping patches and feeds them into a Transformer encoder, utilizing the Attention Mechanism \cite{wang2018self} to extract image features. Features were extracted from ResNet-152 from its final layer. VGG19, a Convolutional Neural Network (Conv-2D)~\cite{cheng2020fast} with 19 layers, employs smaller filters and strides to reduce memory complexity. Features were extracted from the final layer of VGG19 after max-pooling. Faster RCNN, an enhanced version of RCNN~\cite{he2017mask}, is also utilized for feature extraction. Faster-RCNN incorporates a Regional Proposed Network (RPN)~\cite{ye2020sarpnet} on top of the ConvNet-extracted~\cite{tran2017convnet} features to generate object proposals. These proposals were then passed to a classification layer to return bounding boxes. 

\subsection{MML Framework for TM-PathVQA}

\begin{table}[!hbt]
\centering
\caption{Performances of English text-based TM-PathVQA systems for Binary Classification using Top-1 metric (expressed in \%)}
\scalebox{0.93}
{
\begin{tabular}{c|c|c|c}
\hline 
\textbf{Sl. No.} & \textbf{Text Features} & \textbf{Image Features} & \textbf{English} \\ \hline 
1                & LaBSE                  & ViT      &        52.73 \textcolor{blue}{$\downarrow$ (1.65)}         \\ %\hline
2                & LaBSE                  & ResNet-152              &     49.75   \textcolor{blue}{$\downarrow$ (4.63)}          \\ %\hline
3                & LaBSE                  & VGG 19                  &     51.66  \textcolor{blue}{$\downarrow$ (2.75)}           \\ %\hline
4                & LaBSE                  & Faster RCNN             &     \textbf{54.38}             \\ \hline  
\end{tabular}}

\label{tab:binary-text-English}
\end{table}
The proposed MML framework comprises three modules: a feature extraction module for processing input images,  another module for processing input raw audio signals, and a Transformer~\cite{NIPS2017_3f5ee243}  encoder responsible for processing these inputs and generating responses. All these modules form the TM-PathVQA architecture. Speech input sequences were down-sampled using a series of stacked 2-dimensional Convolutional (Conv-2D) layers. Each Conv-2D layer had a stride of $4$ with $5$ channels, resulting in sequence reduction by four times. Then, these audio representations were concatenated with image representations before feeding as input to the Transformer encoder~\cite{NIPS2017_3f5ee243}. \par

The encoder comprises two layers of Transformer blocks, each consisting of two Multi-Head Attentions~\cite{NIPS2017_3f5ee243}  with an output dimension of 64. It is then followed by a Multi-Layer Perceptron (MLP) layer with 128 hidden units. The feed-forward block utilizes 256-dimensional inner states, followed by layer-normalization~\cite{ba2016layer}. A dropout~\cite{srivastava2013improving} rate of 0.2 is applied in the attention block and the MLP layer. The output is then passed to a classification layer represented as $\hat{y} \in \mathbb{R}^{C}$, where $C$ denotes the number of classes (unique words in the text containing answers). For open-ended questions, the number of classes were 4092, while for "Yes/No" type questions, there were only two classes \textit{i.e.} "Yes" or "No." The final classification is then performed using the Softmax activation function.\par
%\subsection{Experimental Settings}
The proposed  MML TM-PathVQA framework was trained for 100 epochs with a batch size of 64. The model with the lowest validation loss was selected for performance evaluation on the test set. Adam \cite{kingma2014adam} optimizer with a Learning Rate (LR) of $1\times10^{-4}$, along with the ReduceLRonPlateau \cite{al2022scheduling} scheduler, was employed. A gradient clipping of 0.1 was also applied. All these experiments were conducted using an A100 GPU with 80 gigabytes (GB) of high-bandwidth memory (HBM2e) and Ubuntu 20.04 LTS as the operating system. \par
\begin{table*}[!hbt]
\centering
\caption{Performance comparison of diverse MML frameworks on TM-PathVQA for multiclass setting across three languages.}
\begin{tabular}{c|c|c|ccccc}
\hline 
\multicolumn{8}{c}{\textcolor{codegreen}{\textbf{ENGLISH}} }                                                                                                                                                                                                                                              \\ \hline 
\multicolumn{1}{c|}{\textbf{Sl. No.}} & \multicolumn{1}{c|}{\textbf{Audio Features}} & \multicolumn{1}{c|}{\textbf{Image Features}} & \multicolumn{1}{c|}{\textbf{Top-1 Accuracy}} & \multicolumn{1}{c|}{\textbf{Bleu-1}} & \multicolumn{1}{c|}{\textbf{Bleu-2}} & \multicolumn{1}{c|}{\textbf{Bleu-3}} & \multicolumn{1}{c}{\textbf{F1-score}} \\ \hline  
1                                 & Wav2Vec2                                 & ViT                       & \multicolumn{1}{c|}{45.76 \textcolor{blue}{$\downarrow$(6.58)} }                 & \multicolumn{1}{c|}{45.95 \textcolor{blue}{$\downarrow$(7.7)}}           & \multicolumn{1}{c|}{4.34 \textcolor{blue}{$\downarrow$(0.33)}}            & \multicolumn{1}{c|}{0.29 \textcolor{blue}{$\downarrow$(0.14)}}            & 2.58 \textcolor{blue}{$\downarrow$(2.73)}              \\ %\hline
2                                 & Wav2Vec2                                 & ResNet-152                               & \multicolumn{1}{c|}{46.39 \textcolor{blue}{$\downarrow$(5.95)}}                  & \multicolumn{1}{c|}{47.02 \textcolor{blue}{$\downarrow$(6.63)}}           & \multicolumn{1}{c|}{3.3 \textcolor{blue}{$\downarrow$(1.37)}}             & \multicolumn{1}{c|}{0.33 \textcolor{blue}{$\downarrow$(0.1)}}            & 4.62 \textcolor{blue}{$\downarrow$(0.69)}              \\ %\hline
3                                 & Wav2Vec2                                 & VGG 19                                   & \multicolumn{1}{c|}{47.79 \textcolor{blue}{$\downarrow$(4.55)}}                  & \multicolumn{1}{c|}{48.05 \textcolor{blue}{$\downarrow$(5.6)}}           & \multicolumn{1}{c|}{4.16 \textcolor{blue}{$\downarrow$(0.51)}}            & \multicolumn{1}{c|}{0.37 \textcolor{blue}{$\downarrow$(0.06)}}            & 4.36 \textcolor{blue}{$\downarrow$(0.95)}             \\ %\hline
4                                 & Wav2Vec2                                 & Faster RCNN                              & \multicolumn{1}{c|}{\textbf{52.34}}         & \multicolumn{1}{c|}{\textbf{53.65}}  & \multicolumn{1}{c|}{\textbf{4.67}}   & \multicolumn{1}{c|}{\textbf{0.43}}   & \textbf{5.31}     \\ \hline
5                                 & Hu-BERT                                  & ViT                       & \multicolumn{1}{c|}{47.96 \red{$\downarrow$ (5.58)}}                  & \multicolumn{1}{c|}{48.81 \red{$\downarrow$ (5.13)}}           & \multicolumn{1}{c|}{3.55 \red{$\downarrow$ (0.67)}}            & \multicolumn{1}{c|}{0.31 \red{$\downarrow$ (0.19)}}            & 4.84\red{$\downarrow$ (0.38)}              \\ %\hline
6                                 & Hu-BERT                                  & ResNet-152                               & \multicolumn{1}{c|}{48.12 \red{$\downarrow$ (5.42)}}                  & \multicolumn{1}{c|}{48.53 \red{$\downarrow$ (5.41)}}           & \multicolumn{1}{c|}{3.71 \red{$\downarrow$ (0.51)}}            & \multicolumn{1}{c|}{0.33 \red{$\downarrow$ (0.17)}}            & 3.3 \red{$\downarrow$ (1.92)}              \\ %\hline
7                                 & Hu-BERT                                  & VGG 19                                   & \multicolumn{1}{c|}{51.63 \red{$\downarrow$ (1.91)}}                  & \multicolumn{1}{c|}{52.17 \red{$\downarrow$ (1.77)}}           & \multicolumn{1}{c|}{3.36 \red{$\downarrow$ (0.86)}}            & \multicolumn{1}{c|}{0.34 \red{$\downarrow$ (0.16)}}            & 4.95  \red{$\downarrow$ (0.27)}            \\ %\hline
8                                 & Hu-BERT                                  & Faster RCNN                              & \multicolumn{1}{c|}{\textbf{53.54}}         & \multicolumn{1}{c|}{\textbf{53.94}}  & \multicolumn{1}{c|}{\textbf{4.22}}   & \multicolumn{1}{c|}{\textbf{0.5}}    & \textbf{5.22}     \\ \hline
9                                 & Mel filterbank                            & ViT                       & \multicolumn{1}{c|}{49.27 \red{$\downarrow$ (2.89)}}                  & \multicolumn{1}{c|}{50.81 \red{$\downarrow$ (1.58)}}           & \multicolumn{1}{c|}{3.44 \red{$\downarrow$ (0.91)}}            & \multicolumn{1}{c|}{0.63 \red{$\downarrow$ (0.04)}}            & 3.92 \red{$\downarrow$ (2.06)}             \\ %\hline
10                                & Mel filterbank                            & ResNet-152                               & \multicolumn{1}{c|}{46.73 \red{$\downarrow$ (4.43)}}                  & \multicolumn{1}{c|}{47.97 \red{$\downarrow$ (4.42)}}           & \multicolumn{1}{c|}{3.36 \red{$\downarrow$ (0.99)}}            & \multicolumn{1}{c|}{0.48 \red{$\downarrow$ (0.19)}}            & 4.5 \red{$\downarrow$ (1.48)}              \\ %\hline
11                                & Mel filterbank                            & VGG 19                                   & \multicolumn{1}{c|}{49.04 \red{$\downarrow$ (3.12)}}                  & \multicolumn{1}{c|}{49.79 \red{$\downarrow$ (2.6)}}           & \multicolumn{1}{c|}{3.06 \red{$\downarrow$ (1.29)}}            & \multicolumn{1}{c|}{\textbf{0.67}}            & 3.5 \red{$\downarrow$ (2.48)}              \\ %\hline
12                                & Mel filterbank                            & Faster RCNN                              & \multicolumn{1}{c|}{\textbf{52.16}}         & \multicolumn{1}{c|}{\textbf{52.39}}  & \multicolumn{1}{c|}{\textbf{4.35}}   & \multicolumn{1}{c|}{0.52 \red{$\downarrow$ (0.15)}}   & \textbf{5.98}     \\ \hline
13                                & Whisper                                  & ViT                       & \multicolumn{1}{c|}{48.08 \red{$\downarrow$ (4.52)}}                  & \multicolumn{1}{c|}{48.74 \red{$\downarrow$ (6.02)}}           & \multicolumn{1}{c|}{3.64 \red{$\downarrow$ (1.35)}}            & \multicolumn{1}{c|}{0.44 \red{$\downarrow$ (0.33)}}            & 4.54 \red{$\downarrow$ (0.65)}             \\ %\hline
14                                & Whisper                                  & ResNet-152                               & \multicolumn{1}{c|}{47.94 \red{$\downarrow$ (4.66)}}                  & \multicolumn{1}{c|}{48.36 \red{$\downarrow$ (6.4)}}           & \multicolumn{1}{c|}{3.54 \red{$\downarrow$ (1.45)}}            & \multicolumn{1}{c|}{0.44 \red{$\downarrow$ (0.33)}}            & 3.11 \red{$\downarrow$ (2.08)}             \\ %\hline
15                                & Whisper                                  & VGG 19                                   & \multicolumn{1}{c|}{50.58 \red{$\downarrow$ (2.02)}}                  & \multicolumn{1}{c|}{51.08 \red{$\downarrow$ (3.68)}}           & \multicolumn{1}{c|}{3.84 \red{$\downarrow$ (1.15)}}            & \multicolumn{1}{c|}{0.51 \red{$\downarrow$ (0.26)}}            & 4.37 \red{$\downarrow$ (0.32)}              \\ %\hline
16                                & Whisper                                  & Faster RCNN                              & \multicolumn{1}{c|}{\textbf{52.6}}          & \multicolumn{1}{c|}{\textbf{54.76}}  & \multicolumn{1}{c|}{\textbf{4.99}}   & \multicolumn{1}{c|}{\textbf{0.77}}   & \textbf{5.19}     \\ \hline 

\multicolumn{8}{c}{\textcolor{codegreen}{\textbf{GERMAN}}}                                                                                                                                                                                                                                              \\ \hline  
\multicolumn{1}{c|}{\textbf{Sl. No.}} & \multicolumn{1}{c|}{\textbf{Audio Features}} & \multicolumn{1}{c|}{\textbf{Image Features}} & \multicolumn{1}{c|}{\textbf{Top-1 Accuracy}} & \multicolumn{1}{c|}{\textbf{Bleu-1}} & \multicolumn{1}{c|}{\textbf{Bleu-2}} & \multicolumn{1}{c|}{\textbf{Bleu-3}} & \multicolumn{1}{c}{\textbf{F1-score}} \\ \hline 
1                                 & Wav2Vec2                                 & ViT                       & \multicolumn{1}{c|}{43.11 \red{$\downarrow$ (5.88)}}                  & \multicolumn{1}{c|}{43.56 \red{$\downarrow$ (6.74)}}           & \multicolumn{1}{c|}{3.56 \red{$\downarrow$ (0.45)}}            & \multicolumn{1}{c|}{0.32 \red{$\downarrow$ (0.18)}}            & 3.27 \red{$\downarrow$ (1.34)}             \\ %\hline
2                                 & Wav2Vec2                                 & ResNet-152                               & \multicolumn{1}{c|}{42.54 \red{$\downarrow$ (6.54)}}                  & \multicolumn{1}{c|}{43.12 \red{$\downarrow$ (7.18)}}           & \multicolumn{1}{c|}{3.34 \red{$\downarrow$ (0.67)}}            & \multicolumn{1}{c|}{0.11 \red{$\downarrow$ (0.39)}}            & 4.11 \red{$\downarrow$ (0.5)}             \\ %\hline
3                                 & Wav2Vec2                                 & VGG 19                                   & \multicolumn{1}{c|}{42.22 \red{$\downarrow$ (6.77)}}                  & \multicolumn{1}{c|}{43.38 \red{$\downarrow$ (6.92)}}           & \multicolumn{1}{c|}{3.21 \red{$\downarrow$ (0.8)}}            & \multicolumn{1}{c|}{0.4 \red{$\downarrow$ (0.10)}}             & 3.32 \red{$\downarrow$ (1.29)}             \\ %\hline
4                                 & Wav2Vec2                                 & Faster RCNN                              & \multicolumn{1}{c|}{\textbf{48.99}}         & \multicolumn{1}{c|}{\textbf{50.3}}   & \multicolumn{1}{c|}{\textbf{4.01}}   & \multicolumn{1}{c|}{\textbf{0.5}}    & \textbf{4.61}     \\ \hline
5                                 & Hu-BERT                                  & ViT                       & \multicolumn{1}{c|}{46.34 \red{$\downarrow$ (7.98)}}                  & \multicolumn{1}{c|}{46.93 \red{$\downarrow$ (7.89)}}           & \multicolumn{1}{c|}{3.04 \red{$\downarrow$ (1.62)}}            & \multicolumn{1}{c|}{0.16 \red{$\downarrow$ (0.57)}}            & 3.38 \red{$\downarrow$ (2.02)}             \\ %\hline
6                                 & Hu-BERT                                  & ResNet-152                               & \multicolumn{1}{c|}{46.88 \red{$\downarrow$ (7.44)}}                  & \multicolumn{1}{c|}{47.18 \red{$\downarrow$ (7.64)}}           & \multicolumn{1}{c|}{3.68 \red{$\downarrow$ (0.98)}}            & \multicolumn{1}{c|}{0.35 \red{$\downarrow$ (0.38)}}            & 4.96 \red{$\downarrow$ (0.44)}             \\ %\hline
7                                 & Hu-BERT                                  & VGG 19                                   & \multicolumn{1}{c|}{50.88 \red{$\downarrow$ (3.44)}}                  & \multicolumn{1}{c|}{51.92 \red{$\downarrow$ (2.9)}}           & \multicolumn{1}{c|}{4.14 \red{$\downarrow$ (0.52)}}            & \multicolumn{1}{c|}{\textbf{0.73}}            & \textbf{5.4}               \\ %\hline
8                                 & Hu-BERT                                  & Faster RCNN                              & \multicolumn{1}{c|}{\textbf{54.32}}         & \multicolumn{1}{c|}{\textbf{54.82}}  & \multicolumn{1}{c|}{\textbf{4.66}}   & \multicolumn{1}{c|}{0.57 \red{$\downarrow$ (0.16)}}   & 4.49 \red{$\downarrow$ (0.91)}     \\ \hline
9                                 & Mel filterbank                            & ViT                       & \multicolumn{1}{c|}{45.87 \red{$\downarrow$ (8.43)}}                  & \multicolumn{1}{c|}{46.56 \red{$\downarrow$ (8.17)}}           & \multicolumn{1}{c|}{3.22 \red{$\downarrow$ (1)}}            & \multicolumn{1}{c|}{0.09 \red{$\downarrow$ (0.32)}}            & 3.21 \red{$\downarrow$ (1.93)}             \\ %\hline
10                                & Mel filterbank                            & ResNet-152                               & \multicolumn{1}{c|}{46.11 \red{$\downarrow$ (8.19)}}                  & \multicolumn{1}{c|}{46.38 \red{$\downarrow$ (8.35)}}           & \multicolumn{1}{c|}{2.28 \red{$\downarrow$ (1.94)}}            & \multicolumn{1}{c|}{0.12 \red{$\downarrow$ (0.29)}}            & 4.54 \red{$\downarrow$ (0.6)}             \\ %\hline
11                                & Mel filterbank                            & VGG 19                                   & \multicolumn{1}{c|}{46.23 \red{$\downarrow$ (8.07)}}                  & \multicolumn{1}{c|}{46.43 \red{$\downarrow$ (8.3)}}           & \multicolumn{1}{c|}{4.11 \red{$\downarrow$ (0.11)}}            & \multicolumn{1}{c|}{0.21 \red{$\downarrow$ (0.2)}}            & 4.84 \red{$\downarrow$ (0.3)}             \\ %\hline
12                                & Mel filterbank                            & Faster RCNN                              & \multicolumn{1}{c|}{\textbf{54.3}}          & \multicolumn{1}{c|}{\textbf{54.73}}  & \multicolumn{1}{c|}{\textbf{4.22}}   & \multicolumn{1}{c|}{\textbf{0.41}}   & \textbf{5.14}     \\ \hline
13                                & Whisper                                  & ViT                       & \multicolumn{1}{c|}{46.46 \red{$\downarrow$ (1.65)}}                  & \multicolumn{1}{c|}{46.76 \red{$\downarrow$ (1.87)}}           & \multicolumn{1}{c|}{4.16 \red{$\downarrow$ (0.54)}}            & \multicolumn{1}{c|}{0.19 \red{$\downarrow$ (0.15)}}            & 4.22 \red{$\downarrow$ (1.01)}             \\ %\hline
14                                & Whisper                                  & ResNet-152                               & \multicolumn{1}{c|}{46.46 \red{$\downarrow$ (1.65)}}                  & \multicolumn{1}{c|}{46.88 \red{$\downarrow$ (1.75)}}           & \multicolumn{1}{c|}{4.02 \red{$\downarrow$ (0.68)}}            & \multicolumn{1}{c|}{0 \red{$\downarrow$ (0.34)}}               & 4.17 \red{$\downarrow$ (1.06)}             \\ %\hline
15                                & Whisper                                  & VGG 19                                   & \multicolumn{1}{c|}{44.78 \red{$\downarrow$ (3.33)}}                  & \multicolumn{1}{c|}{45.27 \red{$\downarrow$ (3.36)}}           & \multicolumn{1}{c|}{4.44 \red{$\downarrow$ (0.26)}}             & \multicolumn{1}{c|}{0 \red{$\downarrow$ (0.34)}}               & 4.2 \red{$\downarrow$ (1.03)}              \\ %\hline
16                                & Whisper                                  & Faster RCNN                              & \multicolumn{1}{c|}{\textbf{48.11}}         & \multicolumn{1}{c|}{\textbf{48.63}}  & \multicolumn{1}{c|}{\textbf{4.7}}   & \multicolumn{1}{c|}{\textbf{0.34}}   & \textbf{5.23}     \\ \hline 

\multicolumn{8}{c}{ \textcolor{codegreen}{\textbf{FRENCH}} }                                                                                                                                                                                                                                              \\ \hline 
\multicolumn{1}{c|}{\textbf{Sl. No.}} & \multicolumn{1}{c|}{\textbf{Audio Features}} & \multicolumn{1}{c|}{\textbf{Image Features}} & \multicolumn{1}{c|}{\textbf{Top-1 Accuracy}} & \multicolumn{1}{c|}{\textbf{Bleu-1}} & \multicolumn{1}{c|}{\textbf{Bleu-2}} & \multicolumn{1}{c|}{\textbf{Bleu-3}} & \multicolumn{1}{c}{\textbf{F1-score}} \\ \hline 
1                                 & Wav2Vec2                                 & ViT                       & \multicolumn{1}{c|}{43.31 \red{$\downarrow$ (5.24)}}                  & \multicolumn{1}{c|}{43.51 \red{$\downarrow$ (5.38)}}           & \multicolumn{1}{c|}{4.8\red{$\downarrow$ (0.09)}}             & \multicolumn{1}{c|}{0.19 \red{$\downarrow$ (0.06)}}            & 4.8 \red{$\downarrow$ (0.4)}              \\ %\hline
2                                 & Wav2Vec2                                 & ResNet-152                               & \multicolumn{1}{c|}{42.39 \red{$\downarrow$ (6.16)}}                  & \multicolumn{1}{c|}{43.89 \red{$\downarrow$ (5)}}           & \multicolumn{1}{c|}{4.05 \red{$\downarrow$ (0.84)}}            & \multicolumn{1}{c|}{0.12 \red{$\downarrow$ (0.13)}}            & 4.81 \red{$\downarrow$ (0.39)}             \\ %\hline
3                                 & Wav2Vec2                                 & VGG 19                                   & \multicolumn{1}{c|}{42.46 \red{$\downarrow$ (6.09)}}                  & \multicolumn{1}{c|}{43.76 \red{$\downarrow$ (5.13)}}           & \multicolumn{1}{c|}{4.66 \red{$\downarrow$ (0.23)}}            & \multicolumn{1}{c|}{0.13 \red{$\downarrow$ (0.12)}}            & 4.4 \red{$\downarrow$ (0.8)}              \\ %\hline
4                                 & Wav2Vec2                                 & Faster RCNN                              & \multicolumn{1}{c|}{\textbf{48.55}}         & \multicolumn{1}{c|}{\textbf{48.89}}  & \multicolumn{1}{c|}{\textbf{4.89} }   & \multicolumn{1}{c|}{\textbf{0.25}}   & \textbf{5.2}      \\ \hline
5                                 & Hu-BERT                                  & ViT                       & \multicolumn{1}{c|}{47.03 \red{$\downarrow$ (6.44)}}                  & \multicolumn{1}{c|}{48.22 \red{$\downarrow$ (5.56)}}           & \multicolumn{1}{c|}{4.86 \red{$\downarrow$ (0.1)}}            & \multicolumn{1}{c|}{0.42 \red{$\downarrow$ (0.61)}}            & 4.85 \red{$\downarrow$ (1.04)}              \\ %\hline
6                                 & Hu-BERT                                  & ResNet-152                               & \multicolumn{1}{c|}{47.98 \red{$\downarrow$ (5.49)}}                  & \multicolumn{1}{c|}{48.06 \red{$\downarrow$ (5.72)}}           & \multicolumn{1}{c|}{4.87 \red{$\downarrow$ (0.09)}}            & \multicolumn{1}{c|}{0.48 \red{$\downarrow$ (0.55)}}            & 3 \red{$\downarrow$ (2.89)}                \\ %\hline
7                                 & Hu-BERT                                  & VGG 19                                   & \multicolumn{1}{c|}{51.66 \red{$\downarrow$ (1.81)}}                  & \multicolumn{1}{c|}{52.14 \red{$\downarrow$ (1.64)}}           & \multicolumn{1}{c|}{4.77 \red{$\downarrow$ (0.19)}}            & \multicolumn{1}{c|}{0.54 \red{$\downarrow$ (0.49)}}            & 4.65 \red{$\downarrow$ (1.24)}             \\ %\hline
8                                 & Hu-BERT                                  & Faster RCNN                              & \multicolumn{1}{c|}{\textbf{53.47}}         & \multicolumn{1}{c|}{\textbf{53.78}}  & \multicolumn{1}{c|}{\textbf{4.96}}   & \multicolumn{1}{c|}{\textbf{1.03}}   & \textbf{5.89}     \\ \hline
9                                 & Mel filterbank                            & ViT                       & \multicolumn{1}{c|}{46.37 \red{$\downarrow$ (5.37)}}                  & \multicolumn{1}{c|}{46.82 \red{$\downarrow$ (6.37)}}           & \multicolumn{1}{c|}{4.07 \red{$\downarrow$ (0.36)}}            & \multicolumn{1}{c|}{0.46 \red{$\downarrow$ (0.15)}}            & 4.68 \red{$\downarrow$ (0.92)}             \\ %\hline
10                                & Mel filterbank                            & ResNet-152                               & \multicolumn{1}{c|}{48.11 \red{$\downarrow$ (3.63)}}                  & \multicolumn{1}{c|}{48.33 \red{$\downarrow$ (9.86)}}           & \multicolumn{1}{c|}{3.7 \red{$\downarrow$ (1.73)}}             & \multicolumn{1}{c|}{\textbf{0.61}}            & 4.96 \red{$\downarrow$ (0.64)}             \\ %\hline
11                                & Mel filterbank                            & VGG 19                                   & \multicolumn{1}{c|}{47.65 \red{$\downarrow$ (4.09)}}                  & \multicolumn{1}{c|}{48.68 \red{$\downarrow$ (4.51)}}           & \multicolumn{1}{c|}{\textbf{4.43}}            & \multicolumn{1}{c|}{0.39 \red{$\downarrow$ (0.22)}}            & 4.73 \red{$\downarrow$ (0.87)}             \\ %\hline
12                                & Mel filterbank                            & Faster RCNN                              & \multicolumn{1}{c|}{\textbf{51.74}}         & \multicolumn{1}{c|}{\textbf{53.19}}  & \multicolumn{1}{c|}{4.04 \red{$\downarrow$ (0.39)}}   & \multicolumn{1}{c|}{0.46 \red{$\downarrow$ (0.15)}}   & \textbf{5.6}      \\ \hline
13                                & Whisper                                  & ViT                       & \multicolumn{1}{c|}{47.16 \red{$\downarrow$ (2.95)}}                  & \multicolumn{1}{c|}{47.96 \red{$\downarrow$ (2.93)}}           & \multicolumn{1}{c|}{4.76 \red{$\downarrow$ (0.01)}}            & \multicolumn{1}{c|}{\textbf{0.4}}             & 4.16 \red{$\downarrow$ (0.45)}             \\ %\hline
14                                & Whisper                                  & ResNet-152                               & \multicolumn{1}{c|}{47 \red{$\downarrow$ (3.11)}}                     & \multicolumn{1}{c|}{47.4 \red{$\downarrow$ (3.49)}}            & \multicolumn{1}{c|}{4.23 \red{$\downarrow$ (0.53)}}            & \multicolumn{1}{c|}{0.19 \red{$\downarrow$ (0.21)}}            & 4.4 \red{$\downarrow$ (0.21)}              \\ %\hline
15                                & Whisper                                  & VGG 19                                   & \multicolumn{1}{c|}{\textbf{50.11}}         & \multicolumn{1}{c|}{\textbf{50.89}}  & \multicolumn{1}{c|}{\textbf{4.77}}   & \multicolumn{1}{c|}{0.21 \red{$\downarrow$ (0.19)}}   & 4.21 \red{$\downarrow$ (0.4)}     \\ %\hline
16                                & Whisper                                  & Faster RCNN                              & \multicolumn{1}{c|}{46.91 \red{$\downarrow$ (3.2)}}                  & \multicolumn{1}{c|}{47.28 \red{$\downarrow$ (3.61)}}           & \multicolumn{1}{c|}{4.19  \red{$\downarrow$ (0.58)}}            & \multicolumn{1}{c|}{0.36 \red{$\downarrow$ (0.04)}}            & \textbf{4.61}              \\ \hline 

\end{tabular}

\label{tab:English-audio-multiclass}
\end{table*}
\subsection{Evaluation Metrics}

The metric used for binary "Yes/No" type questions is Top-1 \cite{niu2012top} accuracy, while for multiclass classification, along with Top-1 accuracy, BLEU-1, BLEU-2, BLEU-3,~\cite{Papineni} and F1 \cite{yacouby2020probabilistic} accuracy are used. Top-1 Accuracy measures the percentage of questions for which the correct answer is ranked first, providing a straightforward assessment of overall accuracy. BLEU-1, BLEU-2 and BLEU-3 metrics were applied to evaluate answer quality based on n-gram overlap, offering insights into exact and partial matches.  F1 score provides a more detailed understanding by considering both false positives and false negatives in handling diverse and open-ended queries about images.

\section{Results \& Discussions}

    Tables \ref{tab:English-text-multiclass}, \ref{tab:English-audio-binary}, \ref{tab:binary-text-English} \& \ref{tab:English-audio-multiclass} presents the performance comparison between various MML-based VQA systems. The performance differences against the system that has the best combination are highlighted in blue. From all the tables, Faster RCNN frameworks outperformed other image features for binary and multiclass classification tasks. From Tables \ref{tab:English-audio-binary} \& \ref{tab:binary-text-English}, the speech-based VQA systems outperformed their text counterparts for both tasks when the best combinations were considered. For multiclass speech-based systems, all the languages achieved similar performance, with English marginally better. Audio features extracted using Hu-BERT also performed better than other acoustic features. From all these tables, it can be inferred that speech-based TM-PathVQA systems have better potential and utility compared to their text counterparts. 
     
Access to the dataset and corresponding code is provided via the following link: \url{https://github.com/aquorio15/path_vqa.git}

\section{Conclusion}
This paper explored the implementation of a speech-based VQA system by introducing the TM-PathVQA dataset. The contributions of this paper include the creation of the first-ever Textless Multi-lingual Pathological VQA dataset in three diverse languages aimed at advancing multi-lingual VQA modeling research. Additionally, this work established baselines for TM-PathVQA systems implemented using various combinations of audio and image features. Speech-based VQA systems, employing Hu-BERT and Faster RCNN, demonstrated superior performance across the three languages compared to text-based systems. Hence, the proposed TM-PathVQA dataset and the experimental benchmarks presented in this study may serve as valuable insights for future research in spoken VQA. Future endeavors will focus on designing novel attention-based MML frameworks that may perform better than the baseline MML frameworks presented in this paper.

\bibliographystyle{IEEEtran}
\bibliography{mybib}

\end{document}